\documentclass[conference]{IEEEtran}
\IEEEoverridecommandlockouts
\usepackage{cite}
\usepackage{amsmath,amssymb,amsfonts}
\usepackage{algorithmic}
\usepackage{graphicx}
\usepackage{textcomp}
\usepackage{xcolor}
\usepackage{booktabs}
\usepackage{subcaption}
\usepackage{caption}
\usepackage{multirow}
\usepackage{float}
\usepackage{url}

\def\BibTeX{{\rm B\kern-.05em{\sc i\kern-.025em b}\kern-.08em
    T\kern-.1667em\lower.7ex\hbox{E}\kern-.125emX}}
\begin{document}

\title{Gradient Boosting Decision Trees on Medical Diagnosis over Tabular Data

}

\author{\IEEEauthorblockN{A. Yark{\i}n Y{\i}ld{\i}z}
\IEEEauthorblockA{\textit{Department of Electrical and Computer Engineering} \\
\textit{Northeastern
University}\\
Boston, MA, USA \\
yildiz.ay@northeastern.edu}
\and
\IEEEauthorblockN{Asli Kalayci}
\IEEEauthorblockA{\textit{The Business School} \\
\textit{Worcester Polytechnic Institute}\\
Worcester, MA, USA \\
akalayci@wpi.edu}
}

\IEEEaftertitletext{%
\vspace{-0.9\baselineskip} 
\tiny © 2025 IEEE. Personal use of this material is permitted. Permission from IEEE must be obtained for all other uses, in any current or future media, including reprinting/republishing this material for advertising or promotional purposes, creating new collective works, for resale or redistribution to servers or lists, or reuse of any copyrighted component of this work in other works.
}

\maketitle

\begin{abstract}
Medical diagnosis is a crucial task in the medical field, in terms of providing accurate classification and respective treatments. Having near-precise decisions based on correct diagnosis can affect a patient's life itself, and may extremely result in a catastrophe if not classified correctly. Several traditional machine learning (ML), such as support vector machines (SVMs) and logistic regression, and state-of-the-art tabular deep learning (DL) methods, including TabNet and TabTransformer, have been proposed and used over tabular medical datasets. Additionally, due to the superior performances, lower computational costs, and easier optimization over different tasks, ensemble methods have been used in the field more recently. They offer a powerful alternative in terms of providing successful medical decision-making processes in several diagnosis tasks. In this study, we investigated the benefits of ensemble methods, especially the Gradient Boosting Decision Tree (GBDT) algorithms in medical classification tasks over tabular data, focusing on XGBoost, CatBoost, and LightGBM. The experiments demonstrate that GBDT methods outperform traditional ML and deep neural network architectures and have the highest average rank over several benchmark tabular medical diagnosis datasets. Furthermore, they require much less computational power compared to DL models, creating the optimal methodology in terms of high performance and lower complexity.

\end{abstract}


\section{Introduction}
\label{sec:introduction}
Tabular data, a prevalent data type in real world applications, consists of rows representing samples and columns representing features of each sample. This form of data is widely utilized across various fields, including manufacturing, finance and healthcare \cite{shwartz2022tabular}. Given its extensive use in applications dependent on relational databases, tabular data modeling is considered crucial in the machine learning (ML) domain. Various traditional ML models have been employed for tabular data classification, including k-nearest neighbors (KNN), logistic regression, and support vector machines (SVMs). Although having a wide range of implementations and high performances in the past, the success of these models has diminished and initiated advancements in other types of ML architectures, such as deep networks and ensemble models. 

Following the increasing availability of large amounts of datasets and advancements in computing resources and deep learning (DL) architectures, such as convolutional models and recurrent mechanisms, have increased their success in the field. With the growing demand for research in large language models (LLMs), attention-based methods have emerged as the state-of-the-art in many downstream tasks, including image, audio, and textual data analysis. Despite these advances, there has been an ongoing debate about whether deep architectures provide satisfying results for tabular data. Following the recent comparisons \cite{borisov2022deep}, deep architectures have deployed superior performance over traditional ML algorithms, over specific tabular data including credit lines, income information, and cartography. Despite increasing the baseline performances compared to other traditional ML methods, DL architectures still possess some challenges in tabular data classification, due to the heterogeneous environment in this type of data. Originally, textual type of data contains a homogeneous environment that helps DL models to have a better generalization. These architectures can encode pretty meaningful representations in the domain of the data. However, tabular data include a large amount of sparse categorical features that impose challenges in learning a representation among these features even having a good encoding before training. Additionally, the numerical features are usually dense and the correlation between the features is much weaker compared to textual data. Therefore, deep neural networks can encounter some challenges when handling tasks involving tabular data.

Due to these issues mentioned above, there is a need for models that require little to no correlational information between features in order to provide good performances. In this context, ensemble models have started to surpass the deep and traditional ML architectures due to their robustness in sparse environments \cite{mohammed2023comprehensive}. For tasks requiring tabular data, ensemble models offer good accuracy while maintaining computational convenience, whereas deep neural networks struggle due to their complex architectures. More recently, gradient boosting decision trees (GBDTs) have started to show superiority and become state-of-the-art in many downstream tasks, including those requiring tabular data \cite{schmitt2022deep}.

Following the recent application fields, one of the most important environments that utilize tabular data is the medical field, especially several diagnosis tasks using patient records. Here, each patient record can be treated as row entries, and columns have the corresponding attributes of the patients. Such medical datasets inherently contain weakly correlated features, meaning that variables often do not exhibit closely associated fluctuations. Understanding these correlations is essential for model interpretation, especially as models grow more complex with numerous interacting variables. Several ML models are interpretable-by-nature such as logistic regression, KNN, and decision trees \cite{molnar2020interpretable}, \cite{bertsimas2018applied}. In particular, GBDTs provide such interpretable model structures that are crucial in medical diagnosis in terms of providing model transparency and user trust, in contrast to the ``black-box'' DL models in which the decision-making process is non-intuitive and difficult to understand \cite{teng2022survey}. Thus, in the possibility of having sparsity and less correlation between each record, robust GBDT methods have gained high popularity in the field \cite{yuan2019high}, \cite{kumar2021catboost}.

Motivated by these advancements, in this work, we highlight the superiority of the ensemble models, especially the GBDTs, over other traditional ML methods and DL architectures, in medical diagnosis tasks, specifically over tabular data. We compared 5 traditional ML methods and 5 DL models with 4 ensemble models, 3 of them being GBDT models, over 7 different medical datasets. We experimentally showed that GBDTs outperform all of the other methods and have the highest ranks over the medical diagnosis datasets on average. Unlike previous works that provide individual applications focusing on a limited number of disease types \cite{murtirawat2020breast} or certain types of tabular datasets outside the medical field \cite{borisov2022deep} by utilizing certain types of ML algorithms \cite{ilyas2021chronic} or a limited number of comparisons regarding DL and ML methodologies \cite{arik2021tabnet}, our contribution is to provide deep insights into the superiority of the GBDT models over state-of-the-art tabular DL and traditional ML  models, focused on a diverse set of tabular medical diagnosis tasks consisting of different dataset dimensions and size settings. Having evaluated the relative performance of the methods across different clinical settings, we also provide insight into which model is more suitable in such medical cases. Additionally, we provide an analysis of training time consumption between the models, which has crucial implications in the medical field in terms of providing fast and accurate well-informed clinical decisions. Our code is publicly available\footnote{\url{https://github.com/yarkin06/TabularGBDT}}.

\section{Background}
\label{sec:background}
\subsection{Traditional Machine Learning Models}

Various studies have explored the application of ML algorithms in medical diagnosis over tabular data. For example, KNN algorithms are used in many applications due to their simplicity, especially in medical data classification, where the data consist of many samples of medical health \cite{xing2019medical}, when detecting breast cancer \cite{murtirawat2020breast}, and providing diagnosis on diabetes by using a fine-tuned KNN classifier \cite{salem2022fine}. Another example can be logistic regression, that has been widely used in both continuous and categorical data classifications, including the medical diagnosis field \cite{boateng2019review}, specifically in COVID-19 diagnosis \cite{podder2021application}, and on many other major chronic disease diagnoses \cite{nusinovici2020logistic}. On the other hand, SVMs are relatively famous for performing medical diagnosis \cite{dinesh2024medical}, especially when combined with other different ML methods, including several deep neural networks \cite{wu2020intelligent}, \cite{latif2022glioma}. Moreover, decision trees are widely used classifiers for tabular data due to their ease of interpretation. Their hierarchical structure allows for straightforward visualization and understanding of the decision-making process \cite{priyanka2020decision}. For medical diagnosis, decision trees are used for coronary artery disease classification \cite{ghiasi2020decision} and chronic kidney disease diagnosis \cite{ilyas2021chronic}. Their performance can be increased when combined with other deep neural network models, such as training the decision trees using convolutional neural networks (CNN) for COVID-19 diagnosis \cite{yoo2020deep}.

\subsection{Deep Neural Networks}

Recently, several DL models that are designed for tabular learning have started to form the state-of-the-art for tabular data classification \cite{gorishniy2021revisiting}. There is a tremendous success of attention-based architectures \cite{vaswani2017attention}, such as transformers that are applied on several different areas, including medical image analysis \cite{he2023transformers}, and as the LLMs have started to occupy the research industry, they have been widely applied to textual data analysis, including the self-attention methods used in time-series analysis \cite{yildiz2022multivariate}, \cite{brown2020language}. In general, they utilize neural network architectures that leverage DL to reflect the complicated relationships between words in the text-based training data set \cite{thirunavukarasu2023large}. The generative AI-powered application, ChatGPT for instance, is a widely used LLM chatbot that produces text in response to text input \cite{thirunavukarasu2023large}. In particular, the LLM algorithm analyzes the data and the context of the words related to each other and creates a text based on a prompt. While DL techniques have advanced, along with enhanced computational resources and big datasets for training, LLM applications have the potential to augment the work across various sectors, including healthcare, which has begun to emerge. \cite{biswas2023role}. 


Moreover, LLM models have shown impressive ability in the field of medicine. They demonstrated the ability to provide meaningful suggestions for further treatments based on the provided information \cite{cascella2023evaluating}. By combining healthcare data and LLMs, medical diagnosis and treatment can achieve improved precision and efficiency. For example, the diagnostic accuracy of algorithms to identify pathology in medical imaging is studied by \cite{aggarwal2021diagnostic}.

The development of DL algorithms is driven by the availability of medical datasets, which are often challenging to access due to their size and diversity. In other words, the proper acquisition of these datasets is integral for interpreting model performance and accuracy within real-world clinical data. Furthermore, LLM processing employs ethical concerns together with security and privacy due to the inclusion of personal health records during model training \cite{kalayci2024exploring}. These concerns and challenges are broadly justified in the literature in terms of its acceptability \cite{kim2023chatgpt}, downstream accuracy \cite{sayin2024can}, medical ethics \cite{goetz2023unreliable}, \cite{ong2024ethical}. Hence, due to the limitations mentioned above, few experimental studies of DL applications in medicine have been conducted, thus, there is a great demand for explanatory research to validate the the diverse models using medical data. On a larger scale, employing diverse deep neural network architectures can serve as a guiding tool for handling more complex data. 

Following these advances of deep architectures in textual type of data, attention-based methods have started to be developed for techniques applied to tabular data as well. Self-attention-based transformer architectures are used for supervised and semi-supervised learning, named TabTransformer \cite{huang2020tabtransformer}, specifically designed for tabular data. Moreover, a sequential attention architecture is developed for choosing which features to benefit from at each decision step, called TabNet \cite{arik2021tabnet}. Furthermore, a feature selection method is utilized for the problems in neural network estimation by using a procedure based on probabilistic relaxation, namely stochastic gates (STG) \cite{yamada2020feature}, and a novel self and semi-supervised learning framework adapted to tabular data classification, named value imputation and mask estimation (VIME) \cite{yoon2020vime}. 

\subsection{Ensemble Models}

Ensemble learning aims to improve model performance by integrating data from various sources, such as healthcare, and medical data with distinct attributes \cite{rider2013ensemble}. One popular approach is Random Forest, which is based on the decision tree methodology and consists of a combination of parallel tree predictors \cite{breiman2001random}. This method employs feature bagging,  where each tree in the ensemble is built from a sample drawn with replacement from the training set, which is also called a bootstrap sample. Random Forest is widely used for tabular data classification across diverse healthcare domains \cite{aria2021comparison}, including medical diagnosis in detecting Alzheimer's disease \cite{bi2020multimodal}, COVID-19 prediction \cite{gupta2021prediction}, and heart disease monitoring \cite{ali2020smart}.

More recently, gradient-boosting methods have started to become the state-of-the-art in many fields, particularly GBDTs. Unlike other ensemble learning methods, many weak learners are combined to create a single strong learner, with each weak learner representing an individual decision tree. Each tree tries to minimize the error on the previous tree, where all of the trees are connected sequentially. Therefore, the model improves at each step iteratively, resulting in a stronger learner by the end. Some of their application fields include GPS signal reception classification \cite{sun2020gradient}, radar waveform classification \cite{yavuz2024detection}, and credit risk assessments \cite{tian2020credit}. Moreover, they occupy various applications in the medical field including heart disease detection \cite{mahesh2022adaboost} and parkinson's disease progression prediction \cite{nilashi2022predicting}. GBDT comprises three main algorithms that have been occupying the field in recent years, which are XGBoost \cite{chen2016xgboost}, LightGBM \cite{ke2017lightgbm}, and CatBoost \cite{prokhorenkova2018catboost}.

XGBoost is a flexible and portable optimized distributed GBDT framework that provides parallel tree boosting in a wide range of applications. It uses a special regularization that maintains an increased efficiency compared to the other GBDT methods. It has been one of the state-of-the-art GBDT methods, and it has diverse application fields including the medical field in breast cancer classification \cite{mahesh2022performance}, 

LightGBM is another prevalent GBDT model that deploys a leaf-wise growth for tree construction, unlike XGBoost's row-by-row approach. Another difference is that it implements a decision tree algorithm that is a highly optimized histogram-based algorithm, that yields increased efficiency and reduced memory consumption. Besides, LightGBM inherits many of XGBoost's advantages such as optimization, regularization capabilities, and support for parallel training.

CatBoost is the other valuable GBDT methodology designed specifically for handling categorical features. It offers advantages such as fast GPU and multi-GPU supports for training, including a range of visualization tools, and use of ordered boosting to overcome overfitting. It is one of the most used ML frameworks in practice, together with XGBoost and LightGBM.

Although the implementations of these algorithms can vary in detail in many applications, their performances usually do not differ that much \cite{prokhorenkova2018catboost}.

\section{Problem Formulation}
\label{sec:problem_formulation}
In general, tabular data consist of multiple attributes that offer information regarding the specific task. Each attribute resides in a separate column in the dataset, and each row corresponds to the different samples. In our task, each sample corresponds to a single patient, and each attribute includes various pieces of information regarding the patient's records relevant to the task at hand.

Formally, for each of the task, we have the feature matrix, that consists of:

\begin{align}
    \boldsymbol{X} = \{\boldsymbol{x}_i\}_{i=1}^n = \{\boldsymbol{x}_1, \boldsymbol{x}_2, \dots, \boldsymbol{x}_n\}
\end{align}

where each $\boldsymbol{x}_i$ denotes the feature vector for the $i$-th patient. Additionally, we have the class labels, that consists of:

\begin{align}
    \boldsymbol{y} = \{y_1, y_2, \dots, y_n \}
\end{align}

where each $y_i$ indicates the diagnosis class for the $i$-th patient. The goal is to learn a mapping function $f: \boldsymbol{X} \rightarrow \boldsymbol{y}$ that can accurately predict the diagnosis class $y_i$ for a given feature vector $\boldsymbol{x}_i$ for a specific patient.

Several techniques can be used in order to overcome possible challenges, including imbalanced datasets and generalization to unseen data, that will be mentioned in Section \ref{sec:experiments}. 

\begin{table}[t]
    \centering
    \caption{Dataset information. The number of features includes the target labels as well.}
    \label{tab:datasets}
    \resizebox{\linewidth}{!}{%
    \begin{tabular}{lccccc}
        \toprule
        \textbf{Datasets} & \textbf{Samples} & \textbf{Features} & \textbf{Classes} & \textbf{Task} \\
        \midrule
        CD \cite{misc_cd} & 70k & 12 & 2 & Binary \\
        Heart Failure \cite{misc_heart_failure_clinical_records_519} & 299 & 13 & 2 & Binary \\
        Parkinsons \cite{misc_parkinsons_174} & 195 & 23 & 2 & Binary \\
        EEG Eye State \cite{misc_eeg_eye_state_264} & 15k & 15 & 2 & Binary \\
        Eye Movements \cite{misc_eye_movements} & 11k & 28 & 3 & Multi-Class \\
        Arcene \cite{misc_arcene_167} & 200 & 10k & 2 & Binary \\
        Prostate \cite{misc_prostate} & 102 & 12.6k & 2 & Binary \\
        \bottomrule
    \end{tabular}
    }
\end{table}

\section{Experiments}
\label{sec:experiments}
\subsection{Datasets}

We evaluated the performance of GBDT models across seven medical diagnosis datasets. Table \ref{tab:datasets} provides details such as the number of samples, features, classes, and task types. The Cardiovascular Disease (CD) dataset \cite{misc_cd} includes 70,000 patient records with 11 features and a binary label, aiming to detect early-stage heart disease. The Heart Failure dataset \cite{misc_heart_failure_clinical_records_519} contains 299 patient records with 13 clinical features, used to predict the likelihood of a death event due to heart failure. The Parkinson's dataset \cite{misc_parkinsons_174} includes 195 samples from 31 patients (23 with Parkinson's), each with 22 biomedical voice measurements, for binary classification. The EEG Eye State dataset \cite{misc_eeg_eye_state_264} has 14 EEG-related features and 15,000 samples, labeled as either eye-open or eye-closed. Eye Movements \cite{misc_eye_movements} consists of 27 features (22 used in psychological studies) for classifying answer relevance (irrelevant, relevant, or correct) in multi-class settings. The Arcene dataset \cite{misc_arcene_167} contains 200 samples with about 10,000 continuous features related to protein abundance, for binary classification of cancer presence. Lastly, the Prostate dataset \cite{misc_prostate} includes 102 samples with roughly 12,600 continuous features to predict prostate cancer status.

\begin{table}[t]
\centering
\caption{Optimal Parameters for MLP}
\label{tab:mlp}
\resizebox{\columnwidth}{!}{%
\begin{tabular}{l|cccccccc}
\toprule
\textbf{Parameters} & \textbf{CD} & \textbf{Heart F.} & \textbf{Parkin.} & \textbf{EEG Eye} & \textbf{Eye Mov.} & \textbf{Arcene} & \textbf{Prostate} \\
\midrule
hidden\_dim & 64 & 98 & 100 & 83 & 65 & 80 & 85 \\
n\_layers & 2 & 5 & 5 & 5 & 4 & 3 & 3 \\
learning\_rate & 0.00099 & 0.00050 & 0.00058 & 0.00082 & 0.00073 & 0.00084 & 0.00092 \\
\bottomrule
\end{tabular}
}
\end{table}

\begin{table}[t]
\centering
\caption{Optimal Parameters for STG}
\label{tab:stg}
\resizebox{\columnwidth}{!}{%
\begin{tabular}{l|cccccccc}
\toprule
\textbf{Parameters} & \textbf{CD} & \textbf{Heart F.} & \textbf{Parkin.} & \textbf{EEG Eye} & \textbf{Eye Mov.} & \textbf{Arcene} & \textbf{Prostate} \\
\midrule
learning\_rate & 0.00364 & 0.04298 & 0.09809 & 0.04842 & 0.02477 & 0.04868 & 0.00984 \\
lam & 0.02405 & 0.48410 & 0.45500 & 0.00244 & 0.03157 & 0.00112 & 0.00645 \\
hidden\_dims & [500, 500, 10] & [500, 50, 10] & [500, 50, 10] & [500, 50, 10] & [500, 200, 20] & [500, 400, 20] & [500, 50, 10] \\
\bottomrule
\end{tabular}
}
\end{table}

\begin{table}[t!]
    \centering
    \caption{Optimal Parameters for TabNet}
    \label{tab:tabnet}
    \resizebox{\columnwidth}{!}{%
    \begin{tabular}{l|cccccccc}
        \toprule
        \textbf{Parameters} & \textbf{CD} & \textbf{Heart F.} & \textbf{Parkin.} & \textbf{EEG Eye} & \textbf{Eye Mov.} & \textbf{Arcene} & \textbf{Prostate} \\
        \midrule
n\_d  &   55 &   14 &  32 &  9 &  50 & 38 & 8 \\
n\_steps &  4 &  10 &  5 &  8 &  4 & 3 & 10 \\
gamma    &  1.27 &  1.38 &  1.11 & 1.14 & 1.01 & 1.60 & 1.98 \\
cat\_emb\_dim   &   3 &  1 & 2 & 3 & 1 &  3 &  2 \\
n\_independent  &  2 &   4 &  3 & 2 &  3 & 4 &  2 \\
n\_shared       &    3 &  4 &  3 & 5 &  3 & 1 & 5 \\
momentum & 0.016 & 0.095 & 0.23 & 0.0041 & 0.16 & 0.34 & 0.40  \\
mask\_type & entmax & entmax & entmax & sparsemax & entmax & entmax & entmax \\
        \bottomrule
    \end{tabular}
    }
\end{table}

\begin{table}[t]
\centering
\caption{Optimal Parameters for TabTransformer}
\label{tab:tabtransformer}
\resizebox{\columnwidth}{!}{%
\begin{tabular}{l|cccccccc}
\toprule
\textbf{Parameters} & \textbf{CD} & \textbf{Heart F.} & \textbf{Parkin.} & \textbf{EEG Eye} & \textbf{Eye Mov.} & \textbf{Arcene} & \textbf{Prostate} \\
\midrule
dim & 128 & 64 & 32 & 64 & 64 & 256 & 64 \\
depth & 12 & 2 & 12 & 6 & 12 & 1 & 2 \\
heads & 4 & 2 & 4 & 4 & 8 & 2 & 4 \\
weight\_decay & -3 & -6 & -4 & -6 & -3 & -5 & -6 \\
learning\_rate & -3 & -3 & -3 & -3 & -3 & -3 & -3 \\
dropout & 0.2 & 0.1 & 0 & 0.4 & 0.5 & 0.4 & 0.2 \\
\bottomrule
\end{tabular}
}
\end{table}

\begin{table}[t]
\centering
\caption{Optimal Parameters for VIME}
\label{tab:vime}
\resizebox{\columnwidth}{!}{%
\begin{tabular}{l|cccccccc}
\toprule
\textbf{Parameters} & \textbf{CD} & \textbf{Heart F.} & \textbf{Parkin.} & \textbf{EEG Eye} & \textbf{Eye Mov.} & \textbf{Arcene} & \textbf{Prostate} \\
\midrule
p\_m & 0.7197 & 0.6619 & 0.8315 & 0.5839 & 0.6613 & 0.2758 & 0.7594 \\
alpha & 6.0055 & 0.3545 & 4.1976 & 4.7264 & 2.4494 & 8.6216 & 7.0457 \\
K & 20 & 3 & 2 & 20 & 2 & 5 & 10 \\
beta & 0.2070 & 0.3099 & 0.3812 & 0.1080 & 0.4152 & 2.4450 & 4.9083 \\
\bottomrule
\end{tabular}
}
\end{table}

\subsection{Data Preprocessing}

Data is preprocessed in the same way for each of the models by applying ordinal encoding to the categorical values, and zero mean, unit variance normalization for numerical features. Furthermore, the categorical features are explicitly specified for the deep neural networks requiring further information on these features, for TabNet and TabTransformer, since their functionality depends on learning appropriate embedding of such features.

\subsection{Training Procedure}

{
We compared 5 traditional ML methods, which are SVM, Logistic Regression, KNN, Decision Tree, and Linear Discriminant Analysis (LDA), and 5 DL models, which are Multilayer Perceptron (MLP), STG, TabNet, TabTransformer, and VIME, with 4 ensemble models including Random Forest, 3 of them being GBDT models that are XGBoost, LightGBM, and CatBoost, over the different medical datasets in Table \ref{tab:datasets}.}

We selected ROC AUC score as our comparison metric. ROC AUC score is the area under the ROC curve that summarizes the the classifier performance with different decision thresholds, and it is a plot of true positive rate (TPR) against the false positive rate (FPR). It is a commonly used metric in comparing multiple models due to its drift tolerance in class balances and reliability in imbalanced data. Hence, we utilized ROC AUC score for both the optimization of the hyperparameters, which is explained below at section \ref{sec:hyperparameter}, and the comparison of different models for each dataset.

In order to evaluate the best-performing model for each respective dataset, we employed 8-fold stratified cross-validation. {We performed the cross-validation by splitting the whole data into 8 parts, taking 7 folds as training and 1 fold as validation, and taking the average and standard deviation of the ROC AUC scores after repeating this process 8 times. Shuffling is set to True by keeping the same seed for every dataset and model for fair comparison and reproducibility.} This technique mitigates the unbalanced data factor that may create a bias in the classification performance. Using each part of the data ensures better generalization rather than selecting a specific part of the data.

\subsection{Hyperparameter Optimization}
\label{sec:hyperparameter}

\begin{table*}[t]
    \centering
    \caption{ROC AUC scores for the traditional ML models, state-of-the-art tabular DL methods, and GBDT models on different datasets. Higher is better. The best scores are bolded, and the second bests are underlined.}
    \label{tab:performance_metrics}
    \resizebox{\textwidth}{!}{%
    \begin{tabular}{l|ccccccc|c}
        \toprule
        \textbf{Model} & \textbf{CD} & \textbf{Heart Failure} & \textbf{Parkinsons} & \textbf{EEG Eye State} & \textbf{Eye Movements} & \textbf{Arcene} & \textbf{Prostate} & \textbf{Avg. Rank} \\
        \midrule
               SVM &                              78.715 ± 0.005 &            86.389 ± 0.048 &         88.791 ± 0.068 &            70.752 ± 0.013 &            78.405 ± 0.007 &     87.094 ± 0.043 &        91.419 ± 0.096 & 9.857 \\
Logistic Reg. &                              78.435 ± 0.005 &            87.571 ± 0.051 &         90.875 ± 0.041 &            61.125 ± 0.014 &            71.180 ± 0.009 &     \textbf{95.211 ± 0.031} &       95.089 ± 0.065 & 8.143 \\
               KNN &                             69.611 ± 0.006 &            77.529 ± 0.067 &         96.857 ± 0.023 &             91.185 ± 0.005 &            72.448 ± 0.009 &     90.869 ± 0.065 &       87.822 ± 0.112 & 9.857 \\
      Random Forest &                             77.464 ± 0.005 &            91.233 ± 0.038 &         96.068 ± 0.033 &              \textbf{98.404 ± 0.002} &            87.234 ± 0.007 &     91.153 ± 0.034 &       93.155 ± 0.078 & 6.000 \\
      Decision Tree &                             63.325 ± 0.006 &            71.646 ± 0.051 &         81.287 ± 0.060 &            83.781 ± 0.008 &            70.951 ± 0.009 &     72.037 ± 0.116 &       80.357 ± 0.106 & 12.714 \\
               LDA &                              70.363 ± 0.005 &             87.896 ± 0.053 &         88.609 ± 0.060 &            67.130 ± 0.014 &            71.273 ± 0.010 &     69.927 ± 0.124 &       93.849 ± 0.060  & 10.571 \\
               \midrule
               MLP \cite{mcculloch1943logical} &                              80.090 ± 0.005 &            87.288 ± 0.056 &         97.186 ± 0.022 &            95.513 ± 0.006 &             73.397 ± 0.015 &     93.669 ± 0.042 &       89.881 ± 0.108 & 6.429 \\
               STG \cite{yamada2020feature} &                              79.667 ± 0.004 &            86.241 ± 0.058 &         95.352 ± 0.038 &            84.854 ± 0.011 &            80.780 ± 0.006 &     90.584 ± 0.062 &       94.048 ± 0.094 & 7.857 \\
            TabNet \cite{arik2021tabnet} &                              77.757 ± 0.004 &            \textbf{93.319 ± 0.037} &         \textbf{99.446 ± 0.012} &            62.441 ± 0.040 &            87.673 ± 0.008 &      87.662 ± 0.098 &        66.865 ± 0.205 & 7.429 \\
    TabTransformer \cite{huang2020tabtransformer} &                             71.327 ± 0.123 &            87.642 ± 0.069 &         96.625 ± 0.027 &            79.646 ± 0.039 &            70.534 ± 0.010 &     \underline{94.724 ± 0.051} &       92.956 ± 0.107 & 8.571 \\
              VIME \cite{yoon2020vime} &                              78.882 ± 0.004 &            85.758 ± 0.047 &         98.532 ± 0.016 &            92.473 ± 0.005 &            81.918 ± 0.008 &     91.721 ± 0.070 &       52.679 ± 0.164 & 7.429 \\
              \midrule
           XGBoost \cite{chen2016xgboost} &                              79.745 ± 0.004 &            90.478 ± 0.025 &         97.265 ± 0.023 &             \underline{98.331 ± 0.002} &            \textbf{89.675 ± 0.008} &     89.123 ± 0.047 &       94.940 ± 0.055 & 4.429 \\
          LightGBM \cite{ke2017lightgbm} &                               \underline{80.296 ± 0.004} &            \underline{91.490 ± 0.027} &         \underline{98.623 ± 0.015} &             97.008 ± 0.004 &            \underline{89.059 ± 0.007} &     91.883 ± 0.043 &       \underline{95.486 ± 0.052} & \textbf{2.571} \\
          CatBoost \cite{prokhorenkova2018catboost} &                              \textbf{80.378 ± 0.004} &            91.056 ± 0.034 &         97.740 ± 0.014 &             97.739 ± 0.003 &            88.954 ± 0.006 &      91.396 ± 0.040 &       \textbf{96.379 ± 0.053} & \underline{3.143} \\
        \bottomrule
\end{tabular}
    }
\end{table*}

\begin{table*}[t]
    \centering
    \caption{Different metric scores for highest ranked (LightGBM) model on different datasets.}
    \label{tab:performance_metrics_lightgbm}
    \resizebox{\textwidth}{!}{%
    \begin{tabular}{l|cccccccc}
        \toprule
        \textbf{Metrics} & \textbf{CD} & \textbf{Heart Failure} & \textbf{Parkinsons} & \textbf{EEG Eye State} & \textbf{Eye Movements} & \textbf{Arcene} & \textbf{Prostate} \\
        \midrule
               Accuracy            &                    73.636 ± 0.005 &   85.615 ± 0.045 &   95.375 ± 0.033 &   90.427 ± 0.012 &   71.726 ± 0.012 &    81.500 ± 0.044 &   91.186 ± 0.046 \\
F1 Score &                     73.598 ± 0.005 &    85.296 ± 0.045 &   95.284 ± 0.034 &   90.407 ± 0.007 &   71.766 ± 0.012 &   81.428 ± 0.045 &   91.113 ± 0.046 \\
ROC AUC           &                     80.296 ± 0.004 &   91.490 ± 0.027 &   98.623 ± 0.015 &    97.008 ± 0.004 &   89.059 ± 0.007 &   91.883 ± 0.043 &   95.486 ± 0.052 \\
Precision           &                    75.538 ± 0.005 &   82.814 ± 0.119 &   96.118 ± 0.035 &   90.883 ± 0.009 &   71.848 ± 0.012 &    84.305 ± 0.062 &   91.443 ± 0.092 \\
Recall              &                   69.865 ± 0.007 &   72.917 ± 0.130 &   97.953 ± 0.026 &   87.446 ± 0.010 &   71.726 ± 0.012 &   83.036 ± 0.050 &   91.667 ± 0.083 \\
        \bottomrule
    \end{tabular}
    }
\end{table*}

Each of the models is optimized over each of the respective datasets in terms of selecting the best combination of hyperparameters. For every model, best-resulting hyperparameter settings are taken with respect to the datasets by evaluating the average ROC AUC score of the 8-folds for each different hyperparameter combination. The hyperparameter combination that reaches the highest average ROC AUC score of the cross-validation is considered as the optimal setting. 

Traditional ML algorithms and ensemble models are easier to optimize compared to deep neural networks due to their less complex architectures. { For the deep learning models, each hyperparameter combination setup is trained for 1000 epochs with an early stopping criterion with 100 epochs patience. On average, 36 different combinations are evaluated for each model, in which the best combination is taken as the best performance of the specific model for comparison.} Optimized hyperparameters for the DL architectures with respect to the datasets are shown in Table's \ref{tab:mlp} - \ref{tab:vime}.

We trained the GPU-supported ML and DL models on NVIDIA A100 Tensor Core GPU, where the average training over the folds lasted 342 minutes at most across all datasets.

\subsection{Results}

Overall results are shown in Table \ref{tab:performance_metrics}. Best models are selected by comparing the average ROC AUC scores of the 8-folds over each respective dataset. For the multi-class problem in the Eye Movements dataset, 'one-vs-one' configuration of the ROC AUC scores are reported. Each result also includes the standard deviation between the different fold results. According to the performances, each model is ranked with respect to the results on each of the datasets, and their average ranks are reported at the end of the table. It can be inferred that GBDT consistently outperforms both the traditional ML methods and the state-of-the-art tabular DL models regardless of the dataset sample size. They perform well in both small sized (e.g. Prostate) and large sized (e.g. CD) datasets. Top-3 ranks consist of the three proposed GBDT models, which demonstrate the overall superior performance of the GBDT over the other traditional ML and tabular DL methods in medical diagnosis tasks.

\begin{figure*}[t]
    \centering
    \begin{subfigure}[b]{0.48\textwidth}
        \centering
        \includegraphics[width=\textwidth]{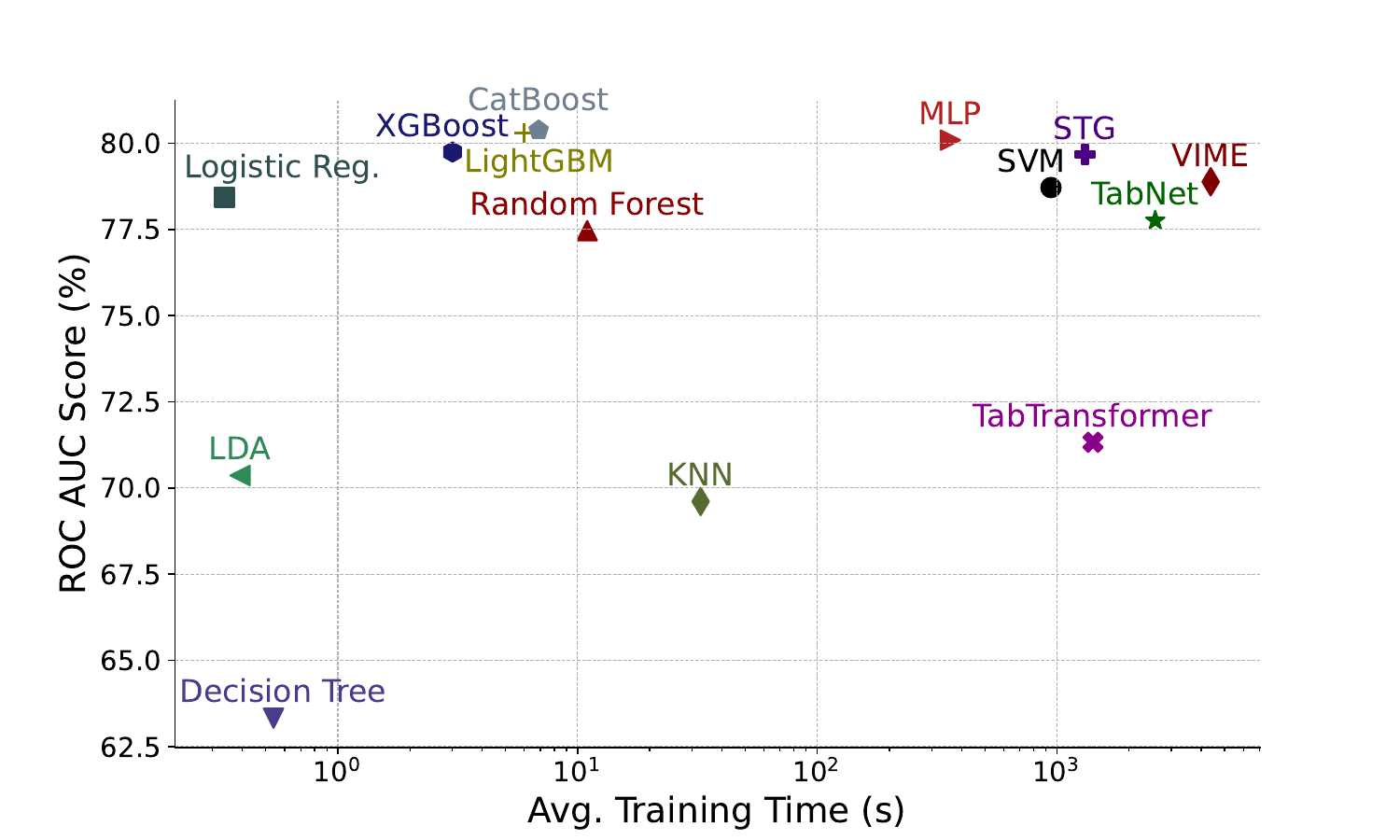}
        \caption{CD dataset}
        \label{fig:cd}
    \end{subfigure}
    \hfill
    \begin{subfigure}[b]{0.48\textwidth}
        \centering
        \includegraphics[width=\textwidth]{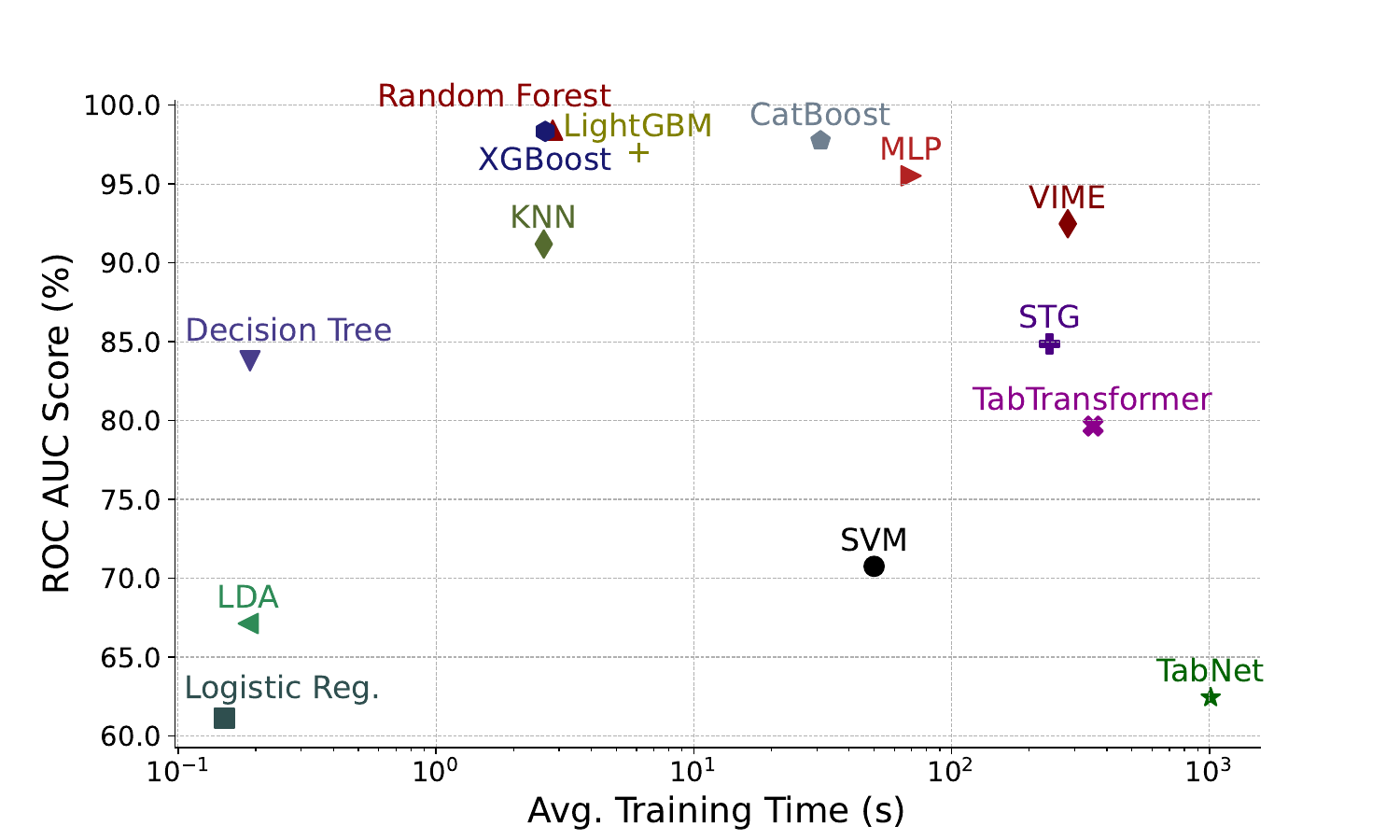}
        \caption{EEG Eye State dataset}
        \label{fig:eeg-eye-state}
    \end{subfigure}
    \hfill
    \caption{ROC AUC score vs avg. training time comparisons for (a) CD and (b) EEG Eye State datasets}
    \label{fig:cd_and_eeg-eye-state}
\end{figure*}

\begin{figure*}[t]
    \centering
    \begin{subfigure}[b]{0.48\textwidth}
        \centering
        \includegraphics[width=\textwidth]{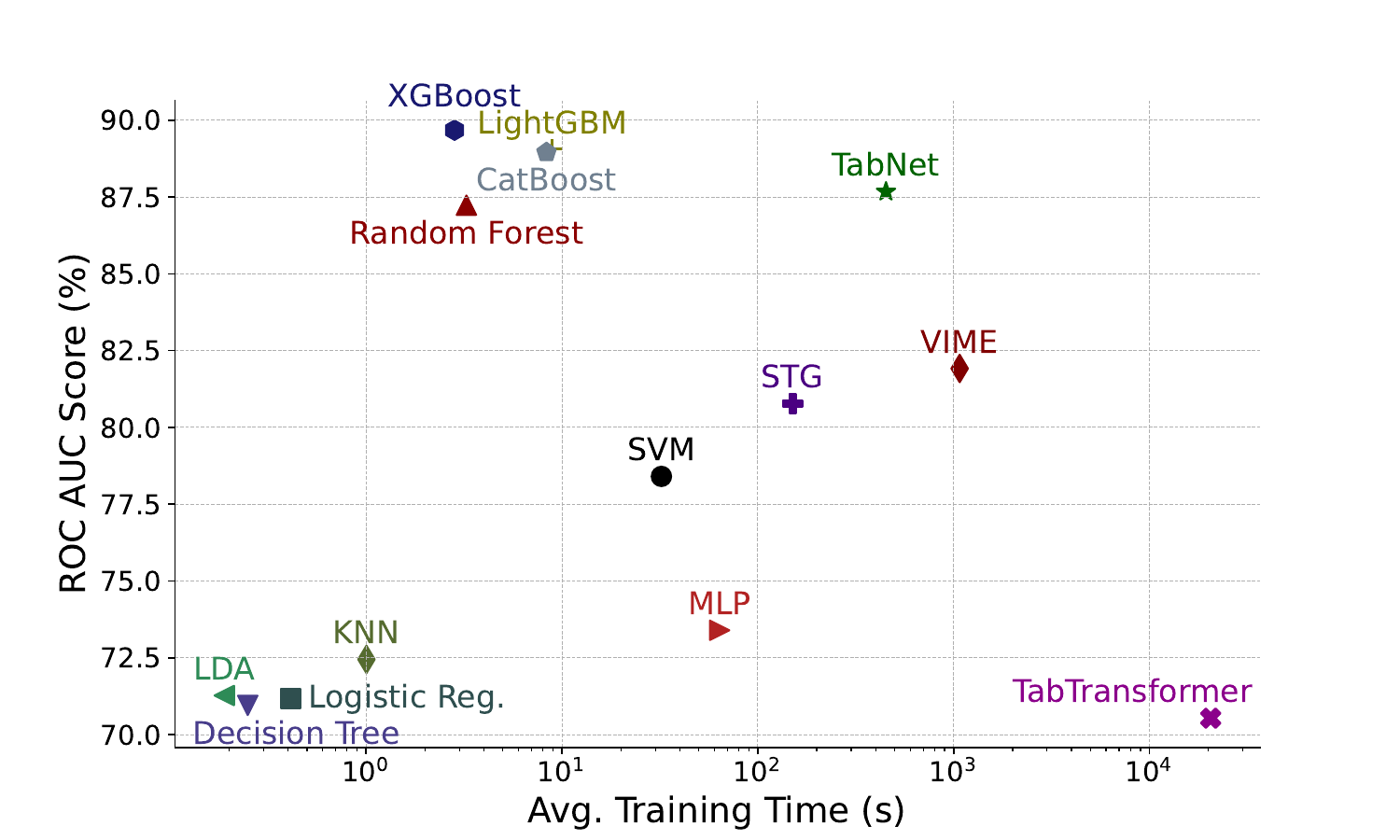}
        \caption{Eye Movements dataset}
        \label{fig:eye_movements}
    \end{subfigure}
    \hfill
    \begin{subfigure}[b]{0.48\textwidth}
        \centering
        \includegraphics[width=\textwidth]{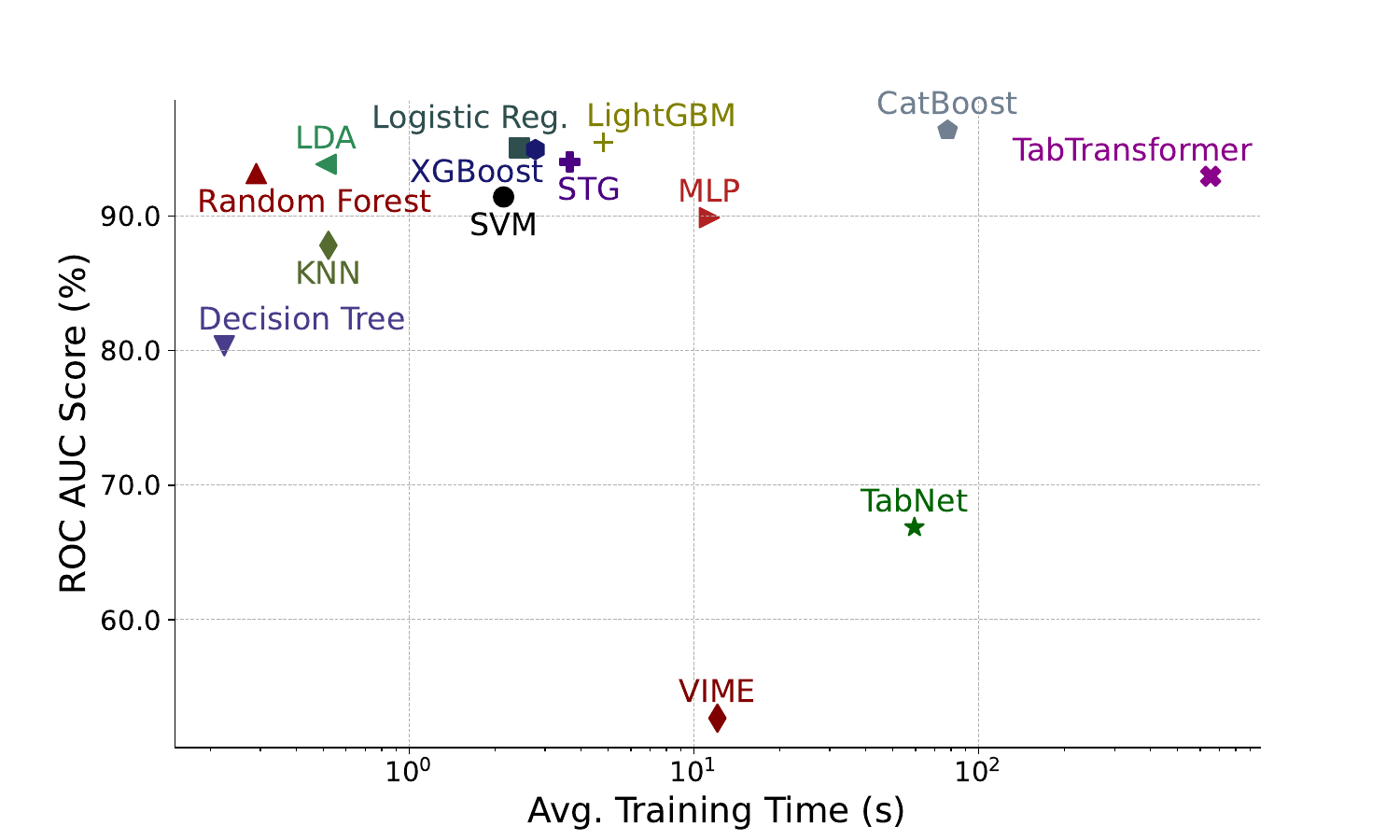}
        \caption{Prostate dataset}
        \label{fig:prostate}
    \end{subfigure}
    \hfill
    \caption{ROC AUC score vs avg. training time comparisons for (a) Eye Movements and (b) Prostate datasets}
    \label{fig:eye_movements_and_prostate}
\end{figure*}

Furthermore, the average ROC AUC scores and average training times of the models are compared. Some examples are visualized over a common plot at Fig.'s \ref{fig:cd_and_eeg-eye-state} and \ref{fig:eye_movements_and_prostate}. Although most tabular DL models perform well on some of the datasets, their average training times are much higher than the training times of GBDT and traditional ML models due to their complex architectures. Having superior performance over the other models, GBDT possesses the optimal structure in terms of performance and time consumption.

Consequently, different metric scores for the highest ranked GBDT model in average, which is LightGBM, are shown in Table \ref{tab:performance_metrics_lightgbm}. Metrics are Accuracy, F1 Score, ROC AUC Score (same as in Table \ref{tab:performance_metrics}), Precision, and Recall. Again, the average and standard deviation of the fold scores are reported. Respective formulas for the metrics are represented at Eq's \eqref{eq:accuracy} - \eqref{eq:precision} except ROC AUC, which is defined in Section \ref{sec:hyperparameter}. \textit{TP}, \textit{FP}, \textit{TN}, and \textit{FN} are defined as true positives, false positives, true negatives, and false negatives respectively. Regarding the multi-class problem in the Eye Movements dataset, weighted averages of the label scores are reported for F1 Score, Recall, and Precision metrics.

\begin{align}
\label{eq:accuracy}
    \text{\textit{Accuracy}} = \frac{TP + FP}{TP + FP + TN + FN}
\end{align}

\begin{align}
\label{eq:f1}
    \text{\textit{F1 Score}} = \frac{2 \times TP}{2 \times TP + FP + FN}
\end{align}

\begin{align}
\label{eq:recall}
    \text{\textit{Recall}} = \frac{TP}{TP+FN}
\end{align}

\begin{align}
\label{eq:precision}
    \text{\textit{Precision}} = \frac{TP}{TP+FP}
\end{align}

Overall, LightGBM provides high performance in medical diagnosis over all of the datasets, yet has a pretty good computational performance by having a relatively less average training time compared to others.

\section{Discussion}
\label{sec:discussion}
Observing Table \ref{tab:performance_metrics}, traditional ML models such as SVM, Logistic Regression, and KNN's generally show varying performance on different tasks, such as KNN excelling on Parkinson's but struggling on others. The reason behind is that SVM and logistic regression are linear models that usually rely on low-dimensional datasets to perform well. Since medical diagnosis datasets usually consist of diverse patient symptoms, such linear models may struggle in performance. On the other hand, being a distance-based method, KNN can perform inconsistently and be sensitive to noisy and high-dimensioned data, which is usually the case for datasets in the medical field. Unlike other ML models, Random Forest provides accurate performances over both small and large sample-sized datasets, such as EEG Eye State and Heart Failure, since it is an ensemble method that becomes effective in reducing overfitting and capturing non-linear relations, being a preferable algorithm for medical datasets with complex patterns. Furthermore, GBDT models, in which we evaluated LightGBM, CatBoost, and XGBoost, tend to perform better on tabular data. These models effectively capture complex interactions between the features that become advantageous for handling imbalanced datasets. In medical settings where the data includes noisy features, GBDT possesses robust mechanisms to overcome these difficulties more easily than other ML methodologies.

In the case of deep neural networks, they tend to have a large number of parameters and complexity compared to other ML-based methods. When the task provides small sample sizes, they can easily overfit the training data and struggle in generalization over the test set, since they tend to capture the noise rather than the correct pattern in the data. Considering MLP and STG, they are mainly designed for tasks including unstructured data, such as images, audio, or text \cite{yamada2020feature}. In contrast to the unstructured types, tabular medical data has fewer correlations between the features, which may potentially include unrelated and redundant features. Therefore, in such tabular tasks, these models tend to have difficulty in performance. Unlike these models; TabNet and VIME have specialized structures for tabular data \cite{arik2021tabnet}, \cite{yoon2020vime}. However, they may still struggle and may not effectively capture the temporal dependencies in several cases, such as in EEG Eye State and Parkinsons, when the features of the tabular data consist of continuous measurements over time rather than having individual entries.

Observing Fig.'s \ref{fig:cd_and_eeg-eye-state} and \ref{fig:eye_movements_and_prostate}, DL models tend to have higher average training times compared to other ML models due to the complex structures that consist of large number of parameters that needs to be optimized. Specifically, in Fig.'s \ref{fig:cd} and \ref{fig:eye_movements}, DL models have higher average training times than traditional ML and GBDT models. Although GBDTs are usually computationally-efficient compared to the DL models, they face challenges in the tradeoff between accuracy and efficiency, mostly in the high feature dimension cases, where the computational complexities become proportional to the number of features \cite{ke2017lightgbm}. As an example, in Fig.'s \ref{fig:eeg-eye-state} and \ref{fig:prostate}, the low-computation trend of GBDTs continues except CatBoost that has increased training time in Fig. \ref{fig:eeg-eye-state} and higher training time than DL models in Fig. \ref{fig:prostate}. The reason behind is that since CatBoost relies on building an ensemble of trees sequentially, due to the existence of high dimensions and intricate patterns in medical datasets, such ensemble models might require deeper and larger number of trees to capture the dependencies in the data, leading to longer training times. Having a relatively higher number of dimensions in the Prostate dataset, CatBoost required a higher average training time than normal, which caused some of the DL models, such as TabNet and VIME, have shorter average training times. In general, GBDT methods employ high overall performance in all of the datasets, yet have a good computational performance by having relatively less average training times.

\section{Conclusion}
\label{sec:conclusion}
In this study, we investigated the overall superiority of ensemble models, especially GBDTs, over other state-of-the-art tabular DL models and traditional ML methods in medical diagnosis, for several benchmark tabular medical datasets. In our analysis, we explored the trade-offs between performance, computational cost, and ease of optimization between the models. Overall, GBDT models exhibited superior performance compared to traditional ML and other deep architectures. Additionally, GBDTs are computationally convenient due to the less complex architectures compared to the deep neural networks that are considerably complex. Due to the less complex structure, GBDTs were much easier to optimize in comparison. Consequently, GBDT models can be safely used in various medical tasks for providing fast and accurate results in any type of diagnosis. {In practice, our findings can facilitate clinical decisions. These insights can help data scientists and medical professionals to optimize model selection based on performance and time efficiency. Within this approach, patient care and healthcare delivery can be improved since our study provides a comprehensive evaluation across a range of ML and DL models as a representative of real-world clinical challenges.}

Despite the significant progress of gradient boosting methods over tabular data, further research can be conducted on benchmarking the advantages of GBDTs across alternative real-life settings in healthcare, as publicly available medical datasets come with limitations such as potential data quality concerns and the static nature of pre-collected data. These alternative settings have the potential to provide further opportunities for analyzing, utilizing, and implementing the GBDT approach. Moreover, they pretty much compose the field of tabular data analysis due to the suffering of deep architectures, that can be extended to any type of task utilized from using tabular data. Due to the significance of tabular data both to the industry and academia, our findings can provide remarkable assistance in the field.

\bibliographystyle{IEEEtran} 
\bibliography{IEEEabrv,refs} 

\vspace{12pt}

\end{document}